\documentclass[conference]{IEEEtran}
\IEEEoverridecommandlockouts

\usepackage{cite}
\usepackage{amsmath,amssymb,amsfonts}
\usepackage{algorithmic}
\usepackage{graphicx}
\usepackage{textcomp}
\usepackage{xcolor}
\usepackage{enumitem}
\usepackage{tablefootnote}
\usepackage{hyperref}
\usepackage[T1]{fontenc}

\def\BibTeX{{\rm B\kern-.05em{\sc i\kern-.025em b}\kern-.08em
    T\kern-.1667em\lower.7ex\hbox{E}\kern-.125emX}}
\begin{document}

\title{CAtCh: Cognitive Assessment through Cookie Thief\\

\thanks{This work was supported in part by Award
Number R01AG066471 from the National Institute Aging of the National
Institutes of Health. This work was also supported in part through the Minerva computational and data resources and staff expertise provided by Scientific Computing and Data at the Icahn School of Medicine at Mount Sinai and supported by the Clinical and Translational Science Awards (CTSA) grant UL1TR004419 from the National Center for Advancing Translational Sciences. Research reported in this publication was also supported by the Office of Research Infrastructure of the National Institutes of Health under award number S10OD026880 and S10OD030463. The content is solely the responsibility of the authors and does not necessarily represent the official views of the National Institutes of Health.}
}

\author{
    Joseph T Colonel\IEEEauthorrefmark{1},
    Carolyn Hagler\IEEEauthorrefmark{2},
    Guiselle Wismer\IEEEauthorrefmark{2},
    Laura Curtis\IEEEauthorrefmark{2},
    Jacqueline Becker\IEEEauthorrefmark{3}, \\
    Juan Wisnivesky\IEEEauthorrefmark{3}, 
    Alex Federman\IEEEauthorrefmark{3},
    Gaurav Pandey\IEEEauthorrefmark{4}\vspace{0.2cm}
    \\
    \IEEEauthorblockA{\IEEEauthorrefmark{1}Windreich Department of AI and Human Health, Icahn School of Medicine at Mount Sinai, New York, NY, USA}
    
    \IEEEauthorblockA{\IEEEauthorrefmark{3}Division of General Internal Medicine, Icahn School of Medicine at Mount Sinai, New York, NY, USA}
    
    \IEEEauthorblockA{\IEEEauthorrefmark{4}Department of Genetics and Genomic Sciences, Icahn School of Medicine at Mount Sinai, New York, NY, USA}

    \IEEEauthorblockA{\IEEEauthorrefmark{2}Division of General Internal Medicine, Feinberg School of Medicine at Northwestern University, Chicago, IL, USA}
}

\maketitle

\begin{abstract}
Several machine learning algorithms have been developed for the prediction of Alzheimer's disease and related dementia (ADRD) from spontaneous speech. However, none of these algorithms have been translated for the prediction of broader cognitive impairment (CI), which in some cases is a precursor and risk factor of ADRD. In this paper, we evaluated several speech-based open-source methods originally proposed for the prediction of ADRD, as well as methods from multimodal sentiment analysis for the task of predicting CI from patient audio recordings. Results demonstrated that multimodal methods outperformed unimodal ones for CI prediction, and that acoustics-based approaches performed better than linguistics-based ones. Specifically, interpretable acoustic features relating to affect and prosody were found to significantly outperform BERT-based linguistic features and interpretable linguistic features, respectively. All the code developed for this study is available at \url{https://github.com/JTColonel/catch}.

\end{abstract}

\begin{IEEEkeywords}
Cognitive impairment, multimodal machine learning, speech processing, natural language processing
\end{IEEEkeywords}

\section{Introduction}
Current estimates indicate that 9\% of all adult Americans and 19\% of those aged 65 years and older have some form of cognitive impairment (CI) \cite{alz_fig,federman2023rates}. Since mild cognitive impairment (MCI) is a risk factor for dementia and because various interventions may reduce the risk for further cognitive decline and/or risk of harm among those with CI, identifying patients with early stage CI may improve long-term management and outcomes for these patients \cite{morley2015brain}. However, MCI and early stage dementia often go undiagnosed, resulting, in part, from low rates of routine CI screening in primary care \cite{yokomizo2014cognitive,amjad2018underdiagnosis}.  

In recent years, researchers have evaluated automated approaches to CI detection to supplant the time consuming task of physician-administered CI screening. A frequent focus of this research has been speech, which is increasingly recognized as a clinical biomarker \cite{kumar2022can,bowden2023systematic,kappen2023speech} of cognitive functioning \cite{de2020artificial}. The proliferation of vocal assistants in smartphones and smart home devices has also brought low cost, high quality voice recording to the market \cite{hoy2018alexa}, greatly expanding the accessibility of this biomarker and the feasibility of speech-based CI screening in clinical settings.   

Several studies that have endeavored to use speech for CI detection have relied on recordings of patients performing the Cookie Thief Test (CTT). The CTT is a 1-minute picture description task used in some neuropsychiatric batteries to assess language and communication abilities and executive functioning in an ecologically valid approximation of spontaneous speech \cite{giles1996performance}. Given the simplicity and brevity of the CTT, it could serve as the basis for a standardized approach to speech-based cognitive impairment screening in clinical settings. 

Early efforts to identify CI using CTT data have been promising. The ADReSSo (Alzheimer's Dementia Recognition through Spontaneous Speech)-2021 Challenge \cite{luz2021detecting} invited participants to predict ADRD from CTT audio recordings. One important aspect of this challenge was that participants were required to use automatic speech recognition (ASR) systems to transcribe the audio to text. ASR systems sidestep the need for labor intensive manual transcriptions, and offer the potential of scaling and automating CI screening from CTT recordings.      

Another relevant field of research for audio-based clinical diagnosis is multimodal sentiment analysis (MSA) \cite{gandhi2023multimodal}. MSA combines text, audio, and video modalities to predict the sentiment of an utterance or sequence of connected utterances. Clinical datasets such as the Distress Analysis Interview Corpus (DAIC) collate recordings of subjects who have been assessed for psychological conditions, such as depression, and had interviews recorded with audio and video \cite{gratch2014distress}. Similar to ML for ADRD classification, automated approaches to psychological assessment have expanded due to the accessibility of the DAIC corpus as well as competitions that have been organized using the data \cite{low2020automated, ringeval2019avec}.  

However, despite the progress made in automated, ML-based detection of ADRD and other neuropsychological conditions, no such methods have been proposed for broader CI prediction from audio recordings, specifically CTT. Furthermore, algorithms proposed for the detection of ADRD have not been systematically translated to the prediction of CI. Although the recent TAUKADIAL Challenge has posed a multilingual audio-based MCI prediction problem, participants had access to substantial additional information in addition to CTT recordings, including demographics, MMSE scores, and fluency tasks that may not be easily available in the clinic \cite{perez2024multilingual}. Thus, the utility of automated methods for predicting CI just from CTT recordings remains unclear. 
\begin{figure*}[ht] 
    \centering
  \includegraphics[width=0.40\linewidth]{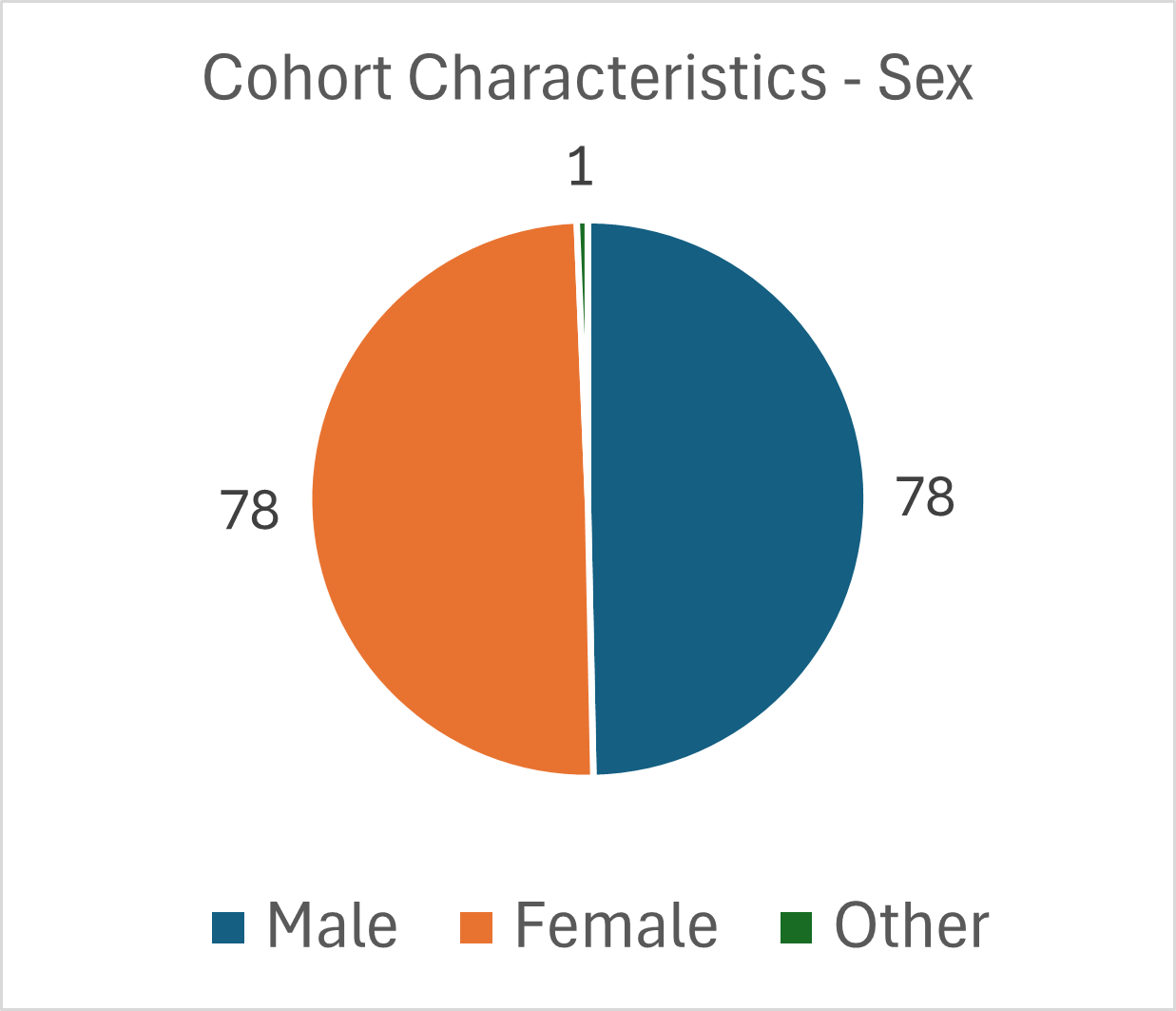}
  \includegraphics[width=0.40\linewidth]{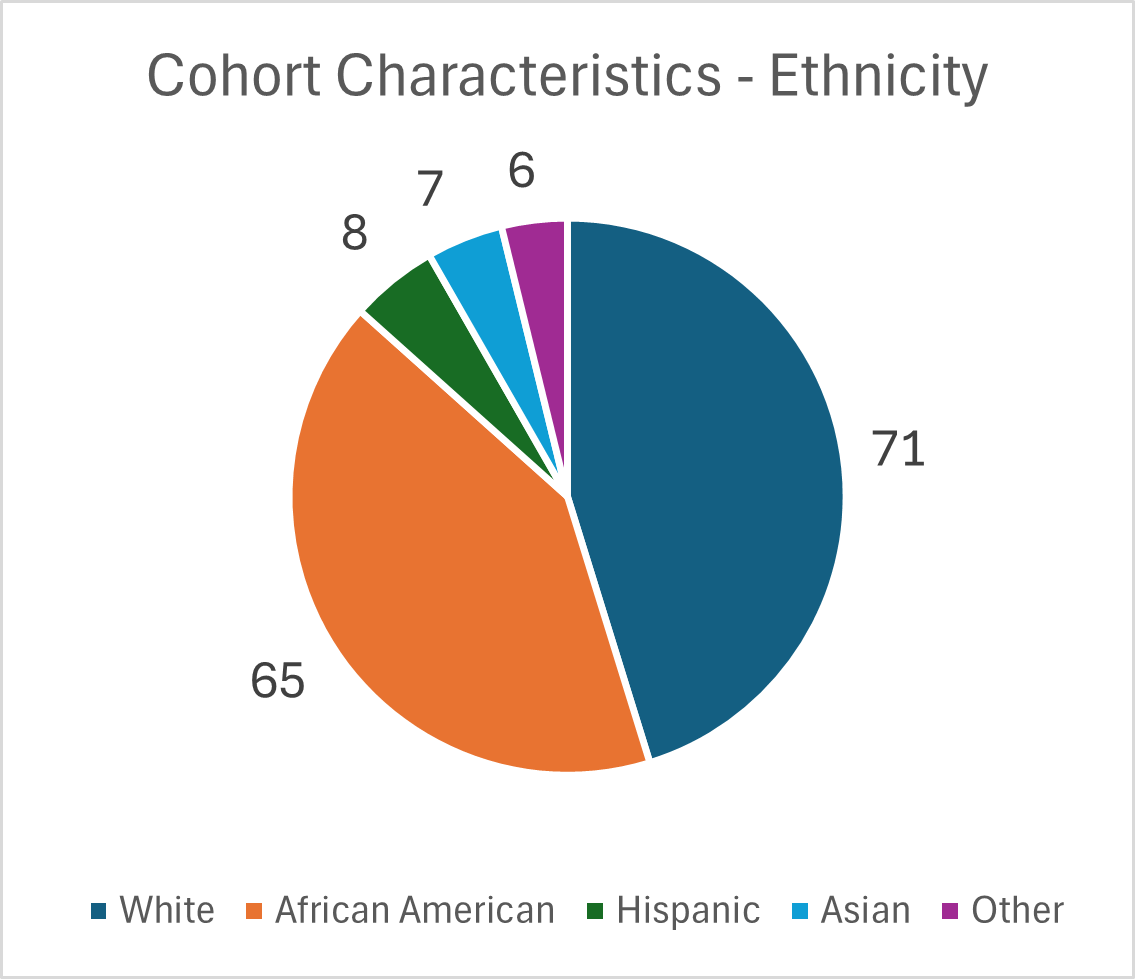}
  \caption {Demographics of the cohort used in this study. \label{demographics}}
\end{figure*}
In this work, we conducted a systemic evaluation of several open-source ML methods for the prediction of CI from CTT recordings. Specifically, the questions we aimed to answer in this study were:  

\begin{enumerate}[label=\textbf{Q\arabic{*}.}]
        \item How well do ML methods developed for MSA and ADRD detection translate to broader CI prediction? 
        \item     What are the general determinants of these methods’ performance when predicting CI? 
\end{enumerate}

This evaluation was conducted on a corpus of CTT recordings collected from a cohort of older adults assessed for CI. As an outcome of this evaluation study, we present a single open-source repository collecting the methods studied in this work, which is expected to spur progress in the automated detection of CI from CTT recordings.  

The rest of the paper is structured as follows. Section II describes the CTT recording corpus used to conduct this evaluation study, including the clinical protocol. Section III outlines the method selection procedure for this study. Section IV describes the evaluation methodology to measure and compare the selected ML methods’ performance on CI prediction. Section V presents the results of the evaluation, with their discussion in Section VI. Limitations and potential future work of the study are described in Section VII.  

\section{The CTT Corpus Used in This Study}

A full description of the clinical study that produced the corpus used in this work can be found in \cite{federman2023rates}. Participants were recruited from primary care practices in Chicago, IL for a study to assess cognitive impairment in primary care patients. They were eligible to participate if their age was 55 years or older, had no diagnosis in their medical record of MCI or dementia, spoke English and did not have a condition that significantly impaired their ability to speak, such as an aphasia. After obtaining informed consent, a research coordinator administered a brief interview consisting of demographic questions and the Montreal Cognitive Assessment (MoCA) \cite{nasreddine2005montreal}. The MoCA is a widely used cognitive screener that consists of 12 tasks covering visuospatial/executive functioning, naming, memory, attention, language, delayed recall, and orientation. Raw MoCA scores were converted to age- and education-adjusted z-scores \cite{rossetti2011normative}. CI was defined as z-scores falling below 1.0 standard deviation of the mean of normative data \cite{rossetti2011normative}. Overall, 28 participants were assessed to have CI and 129 participants were assessed to be cognitively healthy.    

Following the MoCA, the research assistant administered the CTT. Participants were shown an illustration depicting a scene in a kitchen (please refer to Figure 2 in \cite{berube2019stealing}). The research assistant instructed the patient to “Tell me everything you see going on in this picture, as if describing it to someone who is blind” and prompted the participant, if needed, to continue their description for a minimum of 30 seconds and maximum of 60 seconds. Their spoken responses were recorded using a Tascam DR-10L Micro Portable Audio Recorder with Lavalier Microphone attached to a lapel or collar. Of the 200 recruited participants, 157 completed the CTT. 

Transcriptions of the CTT recordings were generated using an extension of OpenAI's Large-V2 Whisper model that can produce word-level timestamps \cite{radford2023robust,lintoai2023whispertimestamped}. Though word-level timestamps are rarely used in ADRD prediction from CTT recordings \cite{braun2022automated}, they are often included in MSA applications \cite{hazarika2020misa,zadeh2018memory}. Voice activity detection was employed to avoid hallucinations in the transcripts \cite{koenecke2024careless}, as well as an initial prompt: “Please separate utterances by period. A dog in the yard. Um, a girl sitting. Mom is washing the dishes.”. A manual review of the 157 transcriptions found no harmful or repetitive hallucinations.  

Overall, a total of 2 hours and 46 minutes of audio were recorded, with an average CTT recording duration of 64.1 seconds (standard deviation 16.1 s). There were 1645 machine-transcribed sentences, averaging 10.5 sentences (sd 4.6) per patient, and 20,257 total words, averaging 129 (sd 45.9) words per patient.

\section{Selection of Methods to Evaluate for CI Prediction}
\label{sec:methods}
\begin{figure}[t] 
    \centering
  \includegraphics[width=\linewidth]{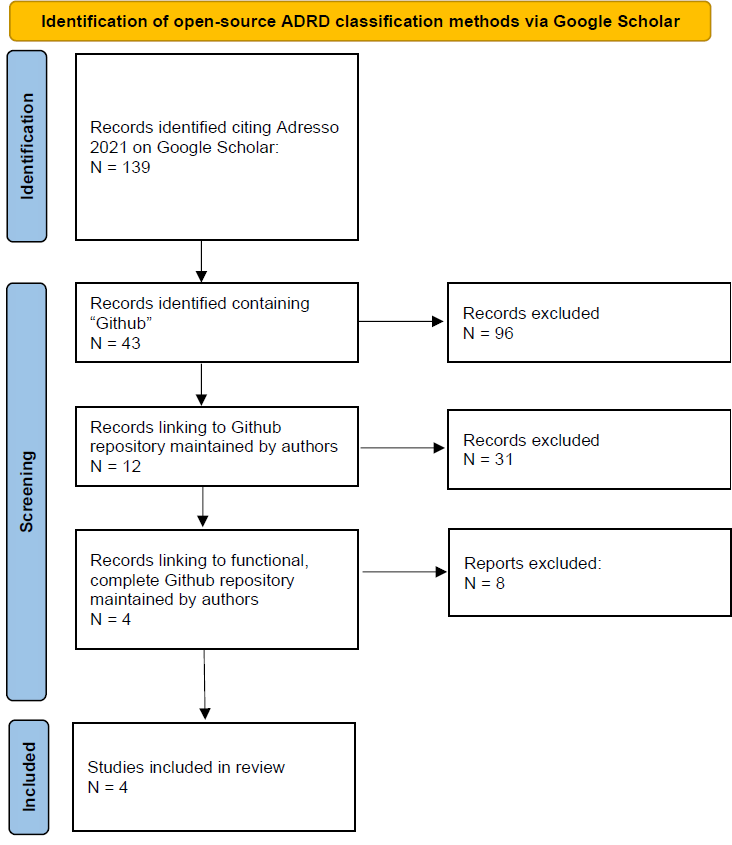}
  \caption {Google Scholar method search performed on August 13, 2024 for this evaluation study \label{method_search}}
\end{figure}
\subsection{ADRD Classification Methods}
We conducted our ADRD-oriented method selection (Figure \ref{method_search}) by searching on Google Scholar on August 13th, 2024 for papers that cited the ADReSSo-2021 Challenge report \cite{luz2021detecting}. This resulted in 139 papers. Next, the additional search term 'github' was used to filter those papers for the ones with open-source code repositories. This resulted in 43 papers. A manual assessment was then conducted on these 43 papers to determine if they contained a link to a github repository of code written by the authors. After this manual assessment, 12 papers included repositories containing code for the methods of the paper. Four of these repositories contained complete, working code that predicted ADRD from recordings of speech and transcripts derived from CTT recordings, which were ultimately included in this study: 
\begin{figure*}[ht] 
  \includegraphics[width=0.49\linewidth]{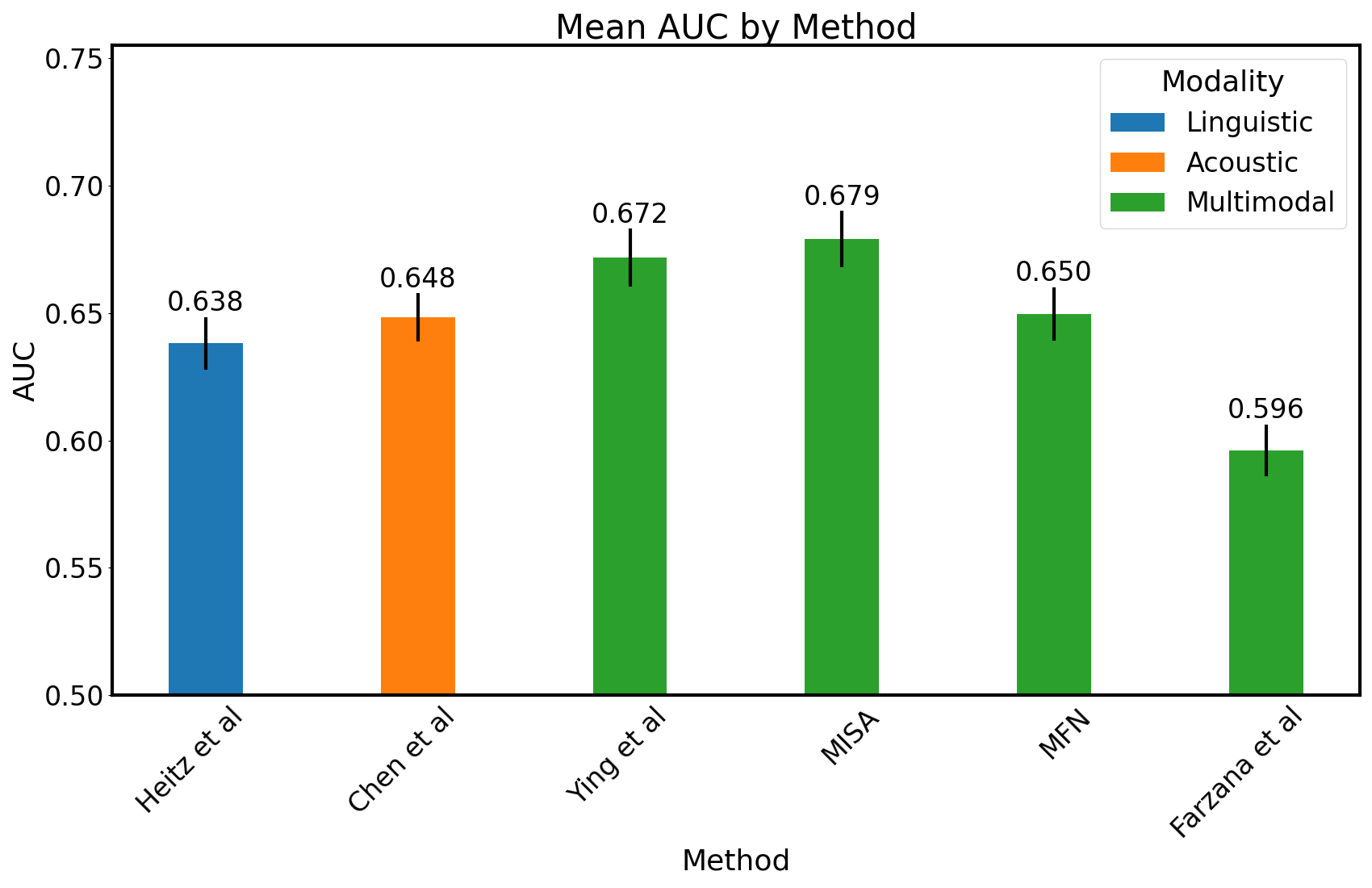} \hfill
  \includegraphics[width=0.49\linewidth]{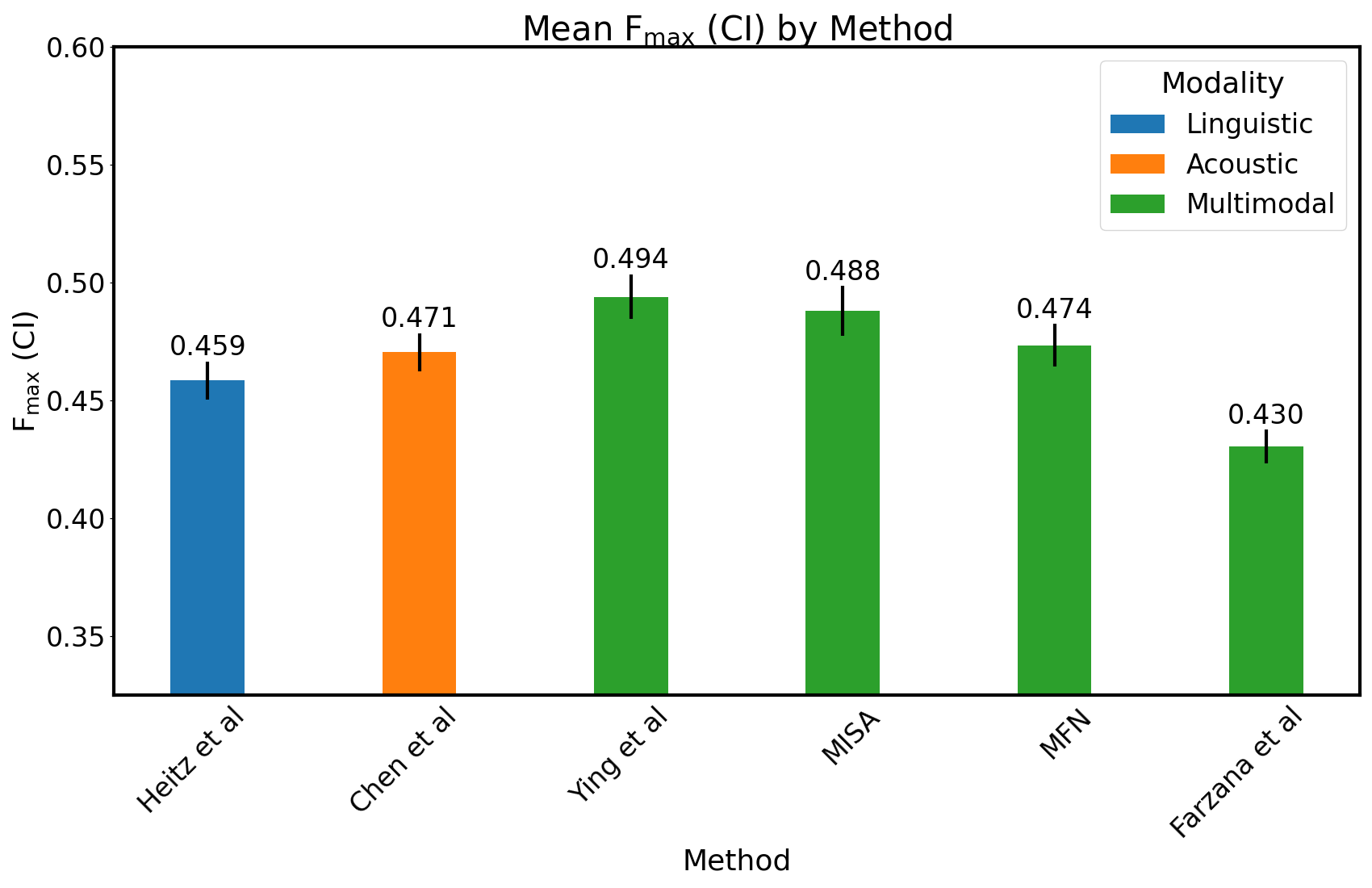}
  \caption {Mean AUC and Fmax (CI) per evaluated method over 100 train-test splits. Error bars show the standard error of the mean. MISA achieved the highest mean AUC, and Ying et al achieved the highest mean Fmax (CI) \label{eval:barcharts}}
\end{figure*}
\subsubsection{Heitz et. al. \cite{heitz2024influence}}
    The authors of this natural language processing (NLP)-based study examined the utility of various ASR systems for ADRD detection from transcripts of CTT recordings. For ASR generated transcripts, the authors found that a random forest classifier with 500 trees trained on expert-defined NLP features performed the best. These features were hand-picked by the authors after searching through the ADRD classification literature, and included syntactic features based on parts-of-speech tags, syntactic features based on grammatical constituents, lexical features, and features of repetitiveness \cite{jurafskyspeech}. 
\subsubsection{Chen et. al. \cite{chenexploring}}
The authors of this acoustics-based method finetuned the Hidden-Unit BERT (HuBERT) model \cite{hsu2021hubert}, which learns acoustic features from speech in a self-supervised manner based on DL clustering techniques, for the prediction of ADRD from CTT recordings. They placed two parallel multilayer perceptron (MLP) classification heads after the pooling layer of HuBERT: one which classifies ADRD, and another which classifies the gender of the speaker. The authors additionally proposed a data augmentation scheme where audio is randomly perturbed, i.e. pitch-shifted, masked in frequency, altered loudness, or vocal tract perturbed. This method was tested in two settings: one in which only the ADReSSo-2021 dataset was used, and another where the ADReSSo-2021 dataset was combined with the Stuttering Events in Podcasts (SEP)-28k dataset \cite{lea2021sep}. In the single-dataset case (used in this work), the authors finetuned the upstream HuBERT model, pretrained on 960 hours of the unlabeled Librispeech dataset \cite{panayotov2015librispeech}, on ten second segments of CTT recordings with a two-second hop length. Predictions of an individual patient's ADRD status were made by averaging the ADRD classifier head's output over all corresponding segments.
\subsubsection{Ying et. al. \cite{ying2023multimodal}}
The authors of this study combined representations from Wav2Vec2.0 (W2V2) \cite{baevski2020wav2vec} and BERT \cite{devlin2018bert}, as well as features from the InterSpeech 2010 Paralinguistic challenge (IS10) \cite{schuller2010interspeech}, all derived from CTT recordings, to predict ADRD. The W2V2 model, pretrained on 960 hours of LibriSpeech data \cite{panayotov2015librispeech}, was finetuned for ADRD classification using the ADReSSo-2021 dataset by appending an MLP classification head to the output of the contextual embeddings. For each patient, five  ten-second segments that uniformly spanned the recording were pooled by the W2V2 contextual embedding, and their respective representations averaged before being passed to the MLP classification head. The BERT model, pretrained on the BookCorpus dataset \cite{zhu2015aligning}, was also finetuned on machine-generated transcriptions of the ADReSSo-2021 dataset, with an MLP classification head applied to the output sequence's CLS classification token. After finetuning of the W2V2 and BERT models, the corresponding representations from the classification heads were extracted, and concatenated with IS10 features to produce one feature vector per patient. The feature vectors were then classified for ADRD status using the default support vector machine (SVM) implementation  with a radial basis kernel \cite{pedregosa2011scikit}.
\subsubsection{Farzana et. al. \cite{farzana2023towards} }
    The authors of this study combined expert-defined NLP and prosody features for multimodal prediction of ADRD. The NLP features included part-of-speech tagging, context-free grammar, syntactic complexity (named entity recognition), vocabulary richness, SUBTL scores \cite{brysbaert2009moving}, and semantic features. The prosody features included duration-based features extracted using the DisVoice toolbox \cite{vasquez2018towards}.  These features were concatenated and then classified using the default  SVM implementation  with a radial basis kernel \cite{pedregosa2011scikit}.  

\subsection{Multimodal Sentiment Analysis (MSA) Methods }
For evaluating MSA methods for CI prediction, we considered the following two prominent architectures: 
\subsubsection{Modality-Invariant and -Specific Representations for MSA (MISA) \cite{hazarika2020misa}}
This method projects unimodal representations of text, video, and speech into a unified latent space using neural network encoders for sentence-level prediction of sentiment. Several loss functions are applied to the distribution of the neural network’s latent space, including modality embedding similarity, difference, reconstruction, and task-specific losses. The linguistic representations are the pooled output of a BERT model, and the acoustic representations are extracted from word-level COVAREP features \cite{degottex2014covarep}. 

In this work, we trained a MISA architecture to predict CI at a patient’s sentence-level text and audio, both derived from their CTT recordings, as input, followed by the averaging of their outputs to predict the final CI status.

\subsubsection{Memory fusion network (MFN) \cite{zadeh2018memory}}
This model consists of a hierarchical system of recurrent neural networks to predict sentence-level sentiment from text, video, and audio modalities. MFN stacks word-level representations using a gated-memory mechanism, and fuses modalities with a classifier head. The linguistic and acoustic representations are derived using Glove \cite{pennington2014glove} and word-level COVAREP features \cite{degottex2014covarep}, respectively.  

Similar to MISA, we trained an MFN architecture to predict CI at the sentence-level given text and audio as input. To predict the CI label for a given patient's CTT, the mean of a patient's sentence-level predictions is taken.

\section{Evaluation Procedure} \label{sec:eval_procedure}
For each CI prediction method described in Section III, we repeated the following evaluation procedure 100 times: 
\begin{enumerate}
    \item Randomly divide the corpus into a 75/25\% train/test split stratified by CI status.
    \item Oversample the minority CI class in the train split to have equal minority/majority representation.
    \item Train the classifier on the training set.
    \item Evaluate the trained classifier on the test set.
\end{enumerate}

For algorithms that used an early-stopping criterion for training neural networks \cite{hazarika2020misa,chenexploring,ying2023multimodal}, the train data was further split 75/25\% to create a train and validation split stratified by CI outcome. Oversampling of the minority CI class only took place on the training split after the creation of the validation split. 

The predictions on the test set from each method were evaluated in terms of the area under the receiver-operating characteristic curve (AUC). Furthermore, to account for the imbalance between the number of CI and non-CI cases in our corpus (28 and 129, respectively), the predictions were also evaluated in terms of the Fmax of the CI class (Fmax  (CI)) \cite{radivojac2013large}, which is the maximum value of the F-measure along the Precision-Recall curve. Means and standard errors were calculated for both AUC and Fmax over the 100 train/test splits as the final evaluation measures.  

To determine whether there was a statistically significant difference in the performance between the evaluated methods, Friedman pairwise testing with a Nemenyi post-hoc correction \cite{demvsar2006statistical} was performed on the measured AUC and Fmax (CI) of each method over the 100 runs.

\section{Results}

\begin{figure*}[h] 
  \includegraphics[width=0.49\linewidth]{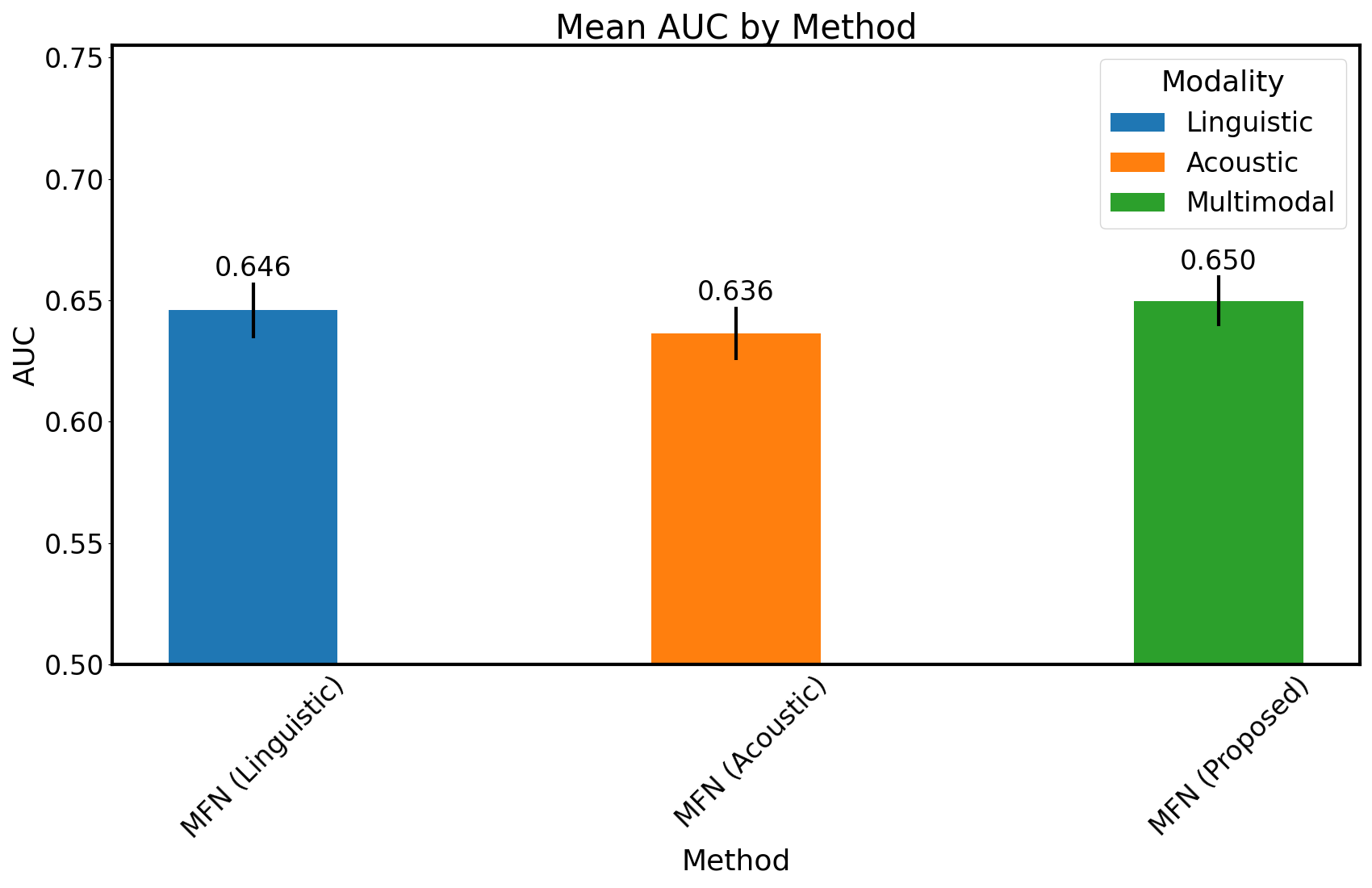} \hfill
  \includegraphics[width=0.49\linewidth]{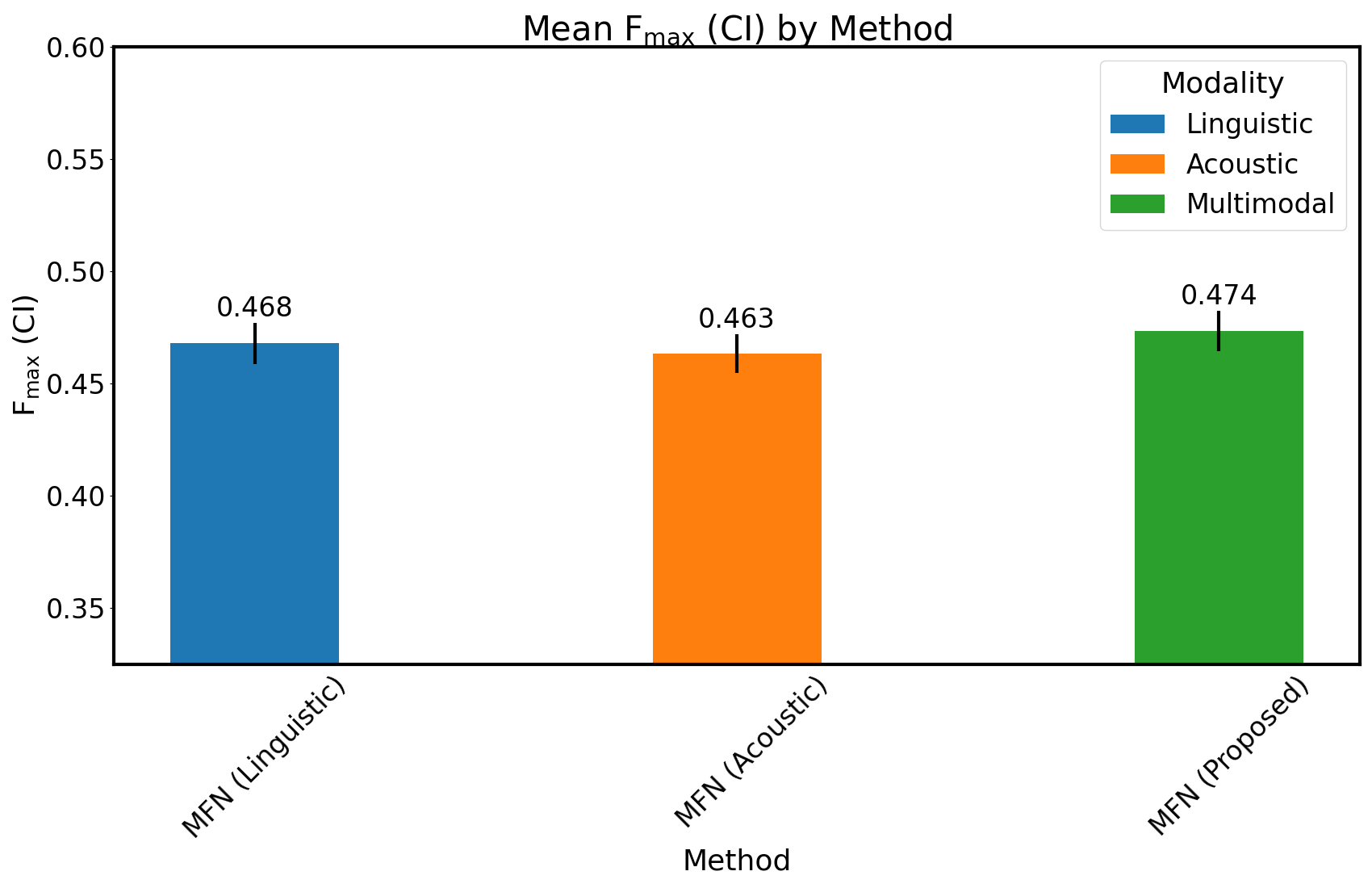}
  \caption {Mean AUC and Fmax (CI) for the MFN classifiers over 100 train-test splits. Error bars represent the standard error of the mean. The proposed multimodal configuration outperforms both unimodal configurations, though no statistically singificant difference was found between the configurations. \label{eval:mfn}}
\end{figure*}

Corresponding to \textbf{Q1} in the Introduction, Figure 3 shows the mean AUC and Fmax (CI) for each prediction method tested in this study. The evaluated methods were able to capture some predictive signal for CI from the CTT recordings used in this study, with significant differences between some of the methods. MISA was the best performer in terms of mean AUC, significantly outperforming Heitz et al (p = 0.027) and Farzana et al (p<0.001). Ying et al was the best performer in terms of mean Fmax (CI), significantly outperforming Heitz et al (p=0.018) and Farzana et al (p<0.001). 

Farzana et al’s AUC performance was significantly lower than MISA (p<0.001), Ying et al (p<0.001), and MFN (p=0.007). In addition, Farzana et al’s Fmax (CI) performance was significantly lower than MISA (p<0.001), Ying et al (p<0.001), MFN (p=0.012), and Chen et al (p=0.013).  

For \textbf{Q2} in the Introduction about the general determinants of CI prediction from the above methods, we  investigated several of them in terms of their constituent parts. To guide our exploration of these results, we structured \textbf{Q2} into the following two sub-questions examining important aspects of these methods: 
To guide our exploration of these results, we propose the following two questions:
\begin{enumerate}[]
        \item \textbf{Q2.a} Do the multimodal formulations of Farzana et. al., MFN, and Ying et. al. outperform their respective unimodal variants? 
        \item \textbf{Q2.b} Does the finetuning of Ying et. al. lead to overfitting and thus a degradation in performance?
        
\end{enumerate}
We note that the construction of MISA breaks down in the unimodal case due to its penalization of private and public encoded representations and thus cannot be evaluated in a unimodal configuration. 

To answer \textbf{Q2.a}, we re-ran MFN and Farzana et al twice: once by ablating the language modality, and once by ablating the acoustic modality. Figures \ref{eval:mfn} and \ref{eval:farzana} show the results of this procedure on MFN and Farzana et al's methods respectively.    

Several permutations of Ying et al's method were run to answer both \textbf{Q2.a} and \textbf{Q2.b}.

\begin{enumerate}
    \item Finetuning the BERT representations, and classifying using only BERT representations.
    \item Not finetuning BERT, and classifying using only BERT representations.
    \item Finetuning the W2V2 representations, and classifying using only W2V2 representations.
    \item Not finetuning W2V2, and classifying using only W2V2 representations.
    \item Classifying using only the IS10 features.
\end{enumerate}

Figure \ref{eval:ying} shows the full results of this procedure for Ying et al's method.   

\section{Discussion}
In this study, we evaluated several open-source ML methods for the prediction of ADRD, as well as open-source methods for MSA, on the task of predicting CI from CTT recordings. As a part of these methods, linguistic and acoustic features were extracted from these speech recordings using both expert-defined methodologies, as well as DL-based architectures. Both traditional ML classifiers and neural network were used to classify these features in unimodal and multimodal configurations. Below, we describe the findings of our study in the form of answers to the guiding questions proposed in the Introduction, as well as other important aspects.     
\subsection{Answering \textbf{Q1}: Do ML methods developed for MSA and ADRD detection translate to CI prediction from CTT recordings?}
Direct comparisons between our evaluated CI prediction methods and those proposed for ADRD prediction in ADReSSo-2021 Challenge due to the Challenge dataset’s 50/50 ADRD/control class balance, which was different from that in our dataset (29/150 CI/control). This difference in class imbalance makes comparison to commonly reported metrics like accuracy difficult. The only comparable metric reported in the ADRD prediction literature selected for this evaluation comes from Heitz et al’s study \cite{heitz2024influence}, which reported an AUC on the ADReSSo-2021 dataset of 0.865. When predicting CI on this cohort, Ying et al’s multimodal method performed best compared to other approaches from the ADRD prediction literature with a mean AUC of 0.672. This suggests that the proposed ADRD prediction methods did not translate sufficiently well to CI prediction. Some degradation in performance is to be expected, as CI encompasses a broad array of outcomes of which ADRD is only a subset. As such, CI may not give rise to the same impairments to the production of spontaneous speech captured by the feature sets used in ADRD detection.  
\begin{figure*}[ht] 
  \includegraphics[width=0.49\linewidth]{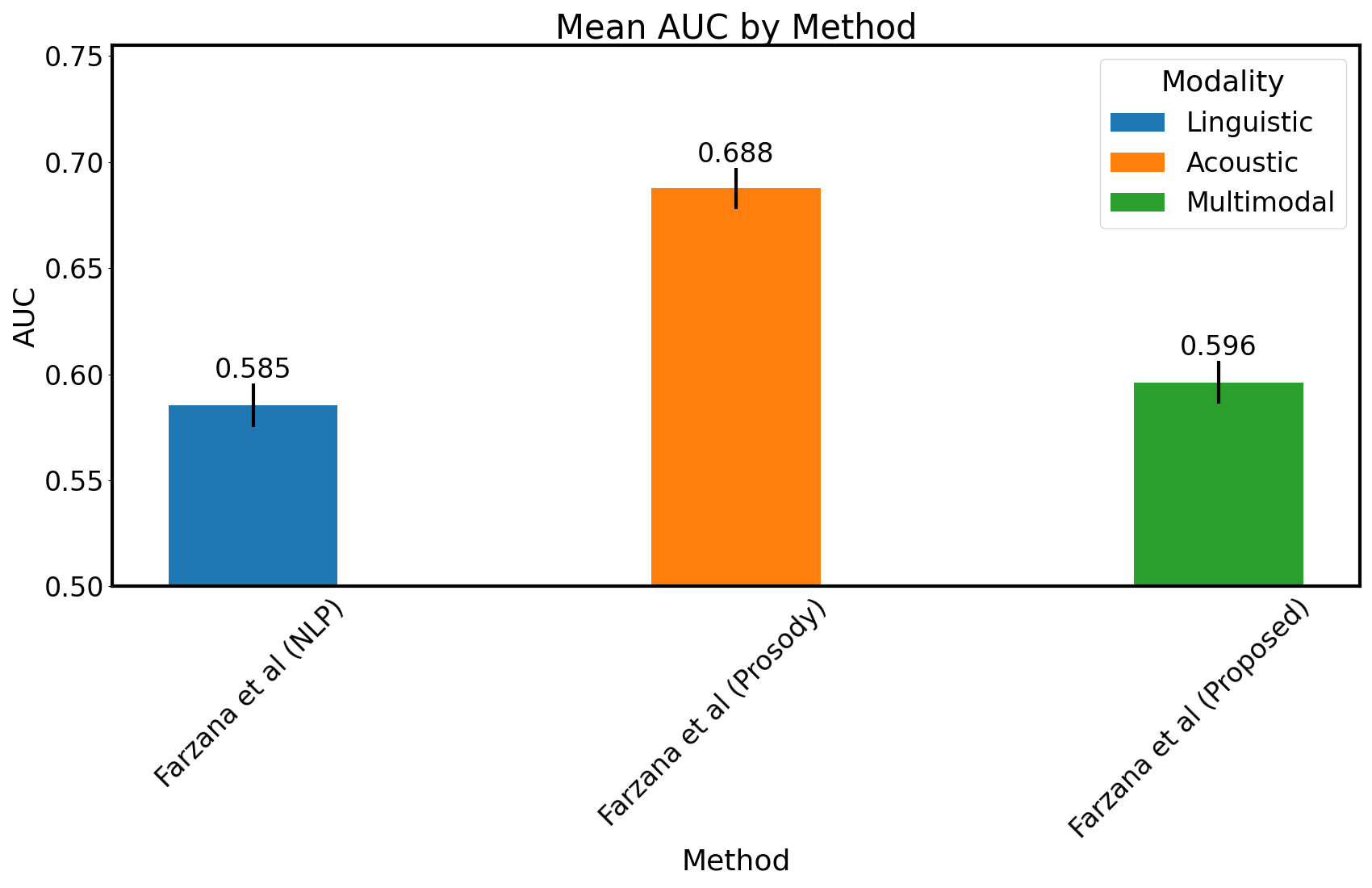} \hfill
  \includegraphics[width=0.49\linewidth]{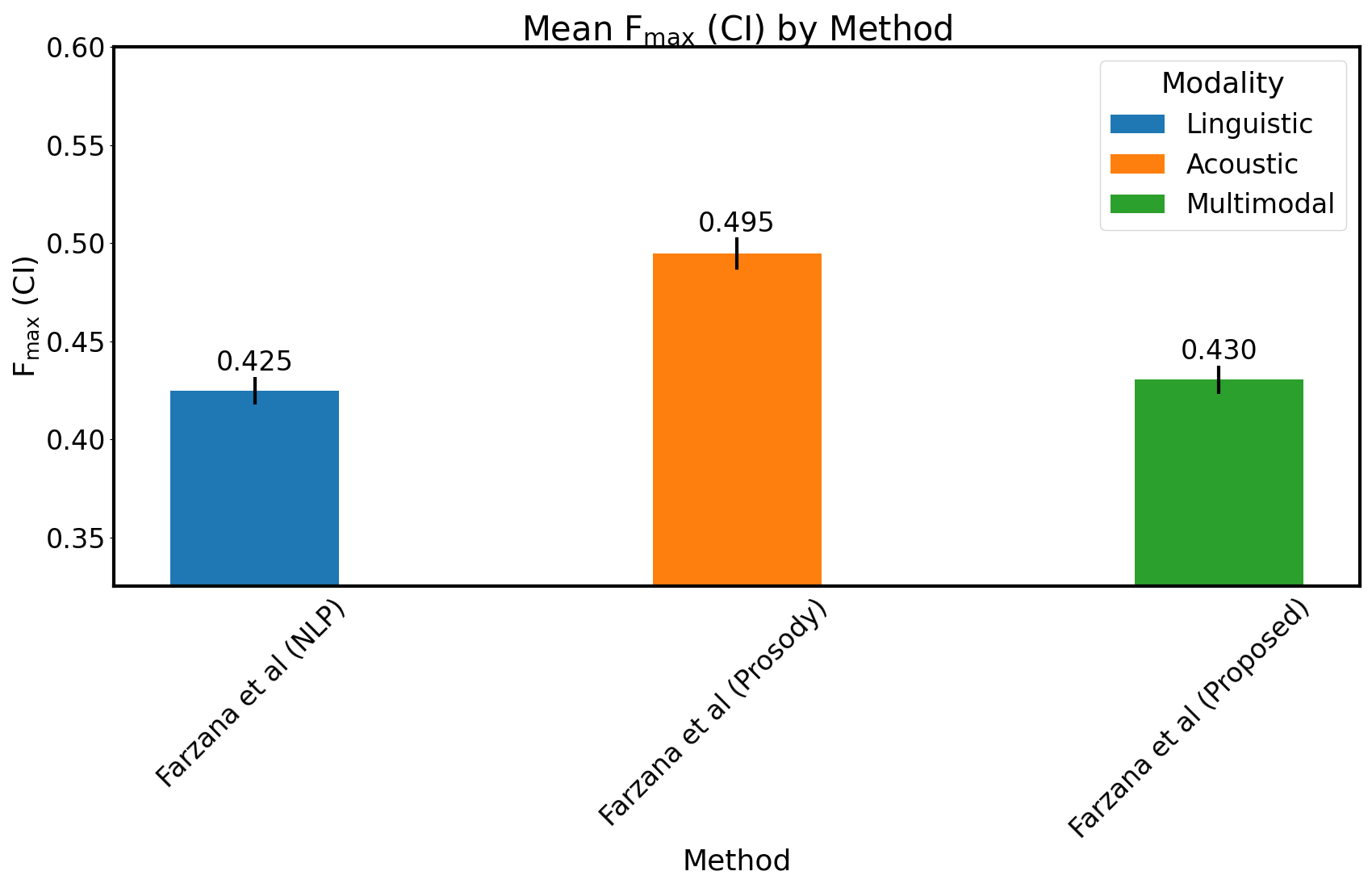}
  \caption {Mean AUC and Fmax (CI) for Farzana et. al. classifiers over 100 train-test splits. Error bars represent the standard error of the mean. The acoustic unimodal configuration achieves the highest median AUC and Fmax (CI) and significantly outperforms the linguistic unimodal configuration (p<0.001) and multimodal configuration (p<0.001). \label{eval:farzana}}
\end{figure*}

Looking at the MSA methods, MISA and MFN performed statistically similarly to Ying et al's method (Figure \ref{eval:barcharts}). Still, in comparison to other uses of MSA for psychological assessment, the performance of these MSA methods was not as good. On the DAIC-WOZ corpus \cite{gratch2014distress}, MISA achieved an F1 score of 0.73, while in this study it achieved an Fmax of 0.488, despite the two datasets having a similar class imbalance and cohort size \cite{jung2024hique}. Thus, while MSA presents an exciting direction for CI prediction, these DL-based methods evaluated here need substantial improvement before their potential is realized. It is possible that with the addition of a video modality, as is present in DAIC-WOZ, as well as larger cohorts, this goal can be accomplished.

\subsection{Answering \textbf{Q2}: What were the general determinants of the evaluated methods’ performance when predicting CI?}
The ADReSSo-2021 Challenge report found that for ADRD prediction, multimodal prediction methods outperformed unimodal methods, and that linguistics-based unimodal methods tended to outperform acoustic-based unimodal methods \cite{de2020artificial}. When these methods were applied to CI prediction on our cohort, however, the trends were not so clear. To clarify these, we broke Q2 into the following two sub-questions focused on general practices in ML/DL-based prediction: 

\subsubsection{\textbf{Q2.a}: Did the multimodal formulations of MFN, Ying et. al., and Farzana et. al. perform better than their respective unimodal variants?}
\phantom{}

In general, the majority of the evaluated multimodal methods outperformed the unimodal ones on our cohort (Figure \ref{eval:barcharts}), though the only significant differences in unimodal vs multimodal performance were observed between MISA and Heitz et al in terms of AUC (p=0.027), Ying et al and Heitz et al in terms of Fmax (CI) (p=0.018), and Farzana et al and Chen et al in terms of Fmax (CI) (p=0.013). Furthermore, Farzana et al’s multimodal method was found to significantly underperform Chen et al’s acoustics-based unimodal method in terms of Fmax (CI) (p=0.013).  

\begin{figure*}[ht] 
  \includegraphics[width=0.49\linewidth]{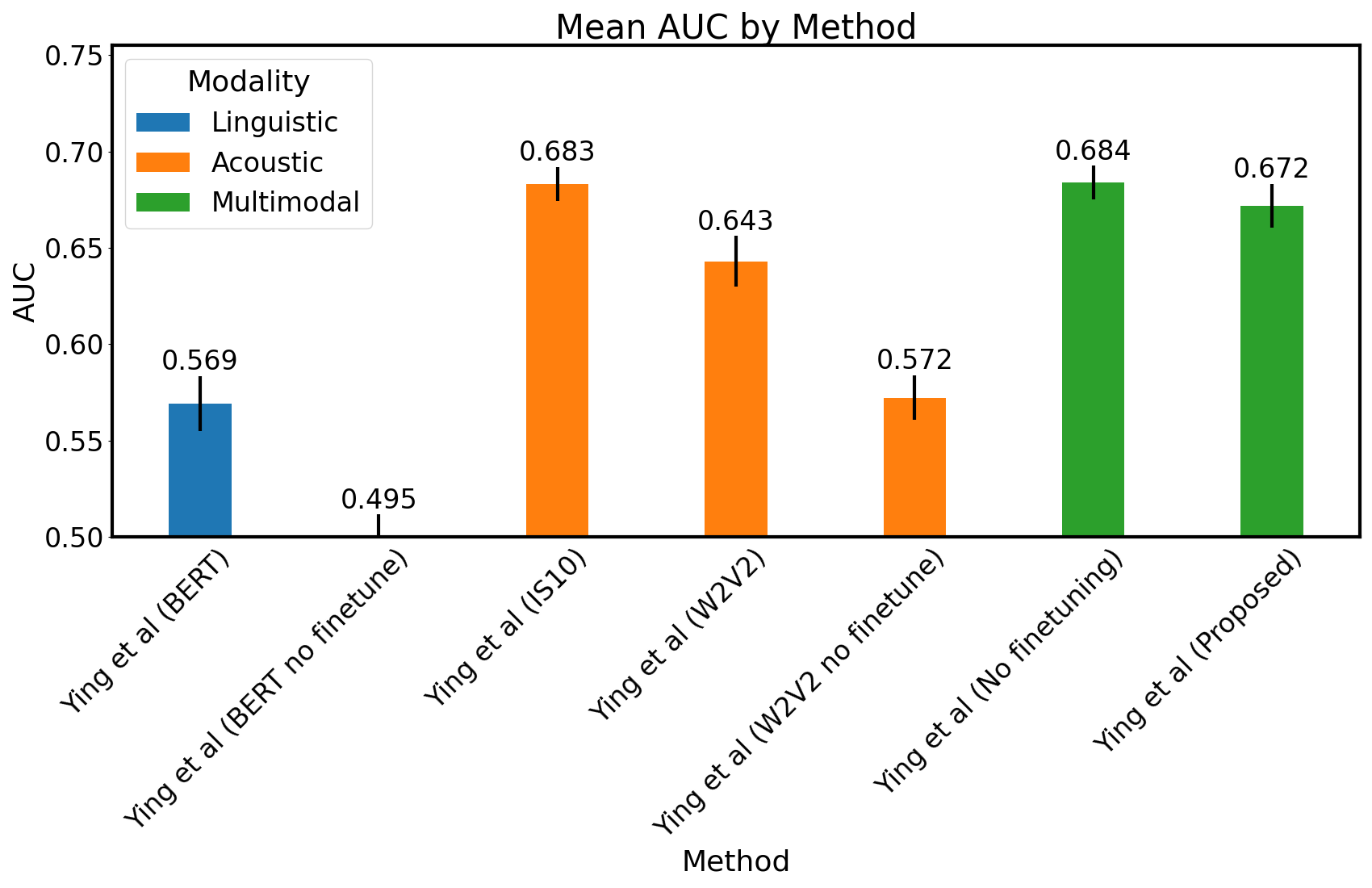} \hfill
  \includegraphics[width=0.49\linewidth]{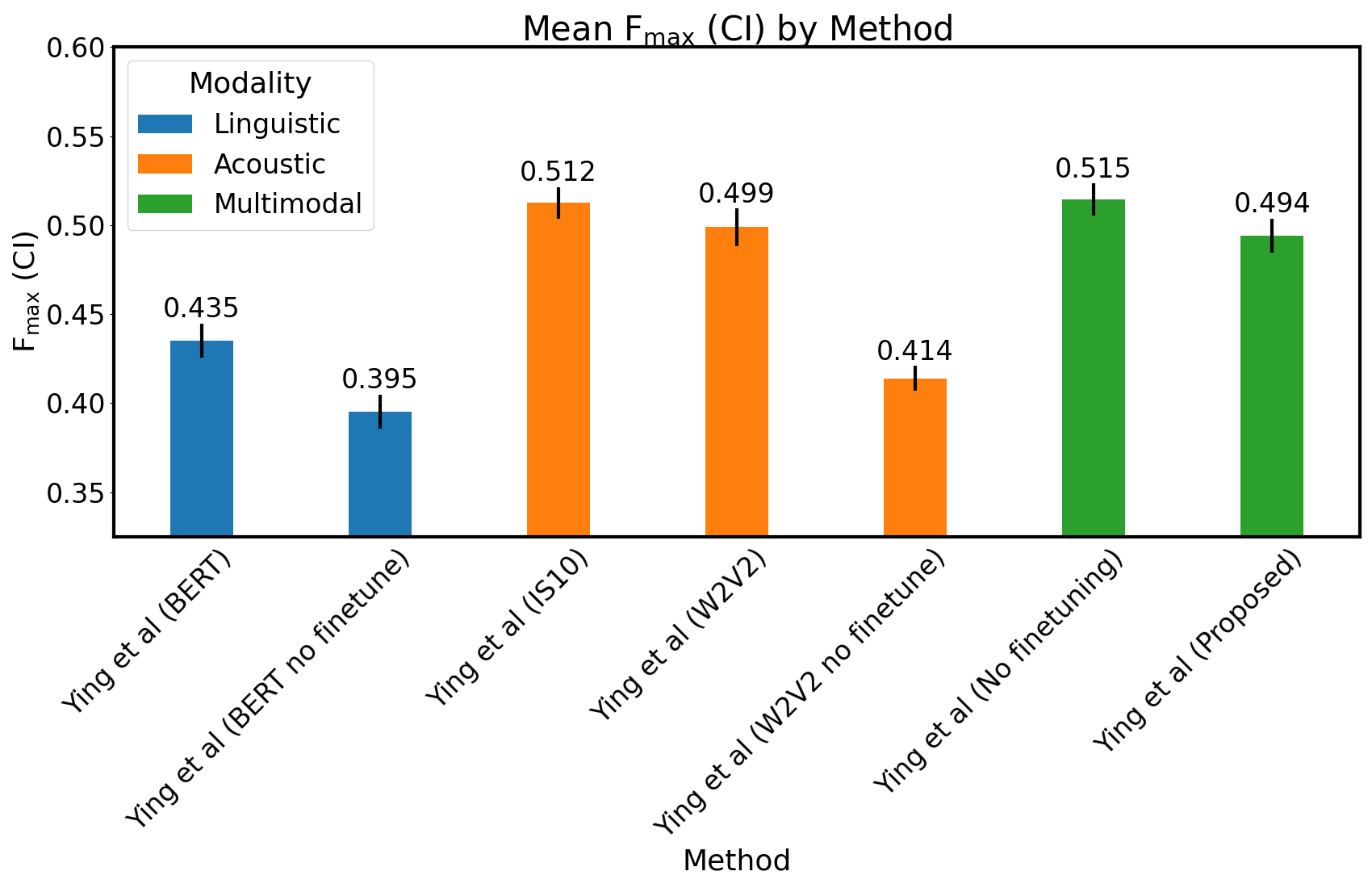}
  \caption {Mean AUC and Fmax(CI) for various Ying et. al. configurations over 100 train-test splits. Error bars represent the standard error of the mean. Ying et. al.'s proposed configuration with no finetuning achieves the highest AUC and Fmax(CI), though this was not found to be a statistically signficant difference from the proposed configuration with finetuning, the IS10-only configuration, and finetuned W2V2 configuration. In both AUC and Fmax(CI), Ying et. al.'s proposed configuration with no finetuning significantly outperformed the finetuned BERT configuration (p<0.001), non-finetuned BERT (p<0.001), and non-finetuned W2V2 configuration (p<0.001). \label{eval:ying}}
\end{figure*}

To further understand these trends, we investigated specific methods in terms of their multimodal configurations and their unimodal components. We focused on MFN, Ying et. al., and Farzana et. al. for this investigation, since the unimodal formulations of MISA break down due to the multimodal penalizations of the latent space. For MFN and Ying et. al,  whose results corresponding to this question are shown in Figures \ref{eval:mfn} and \ref{eval:ying} respectively, the multimodal configurations outperformed the unimodal configurations, as expected \cite{baltruvsaitis2018multimodal}.  However, the difference in performance was slight compared to the best-performing unimodal configurations. For MFN, no difference in performance was found to be significant (p>0.9 for all measures). For Ying et al, the multimodal configurations were found to significantly outperform both BERT-based unimodal classifiers, as well as the non-finetuned W2V2-based classifier (p<0.001 for all measures). 

For Farzana et. al., however, Figure \ref{eval:farzana} shows that the unimodal prosody-based acoustic classifier significantly outperformed the multimodal classifier in terms of both AUC and Fmax (CI) (p<0.001 for both), as well as the linguistics-based classifier (p<0.001 for both). This demonstrates the importance of testing unimodal classifiers alongside their multimodal configurations, as these results show the inclusion of the linguistic modality harms the performance of Farzana et al’s multimodal classifier. 

\subsubsection{\textbf{Q2.b}: Did the finetuning of Ying et al’s model. lead to overfitting and thus a degradation in performance?}

Finetuning large neural networks on small, out-of-distribution datasets has been shown to sometimes distort features and hurt classification performance, though this observation is not a guarantee \cite{kumar2022fine}. We tested the effect of this factor on the performance of Ying et al’s method in our study.  Specifically, we wanted to determine if finetuning the BERT and W2V2 components of this model could lead to overfitting, and thus deteriorate performance on the test set. Chen et al’s method, the only other method to use a pretrained DL architecture, was not considered for this analysis as the multitask representation learning breaks down without finetuning.  

The green bars in Figure \ref{eval:ying} demonstrated that not finetuning the W2V2 and BERT upstream models did yield a slight improvement in mean AUC and Fmax (CI) when the acoustic and linguistic modalities were fused for prediction, though this improvement was not found to be significant (p>0.9 for both measures). For the BERT-based unimodal classifiers, mean AUC and Fmax (CI) improved, though neither improvement was statistically significant (p=0.31 and p=0.07 respectively for the two measures). However, the W2V2-based unimodal classifiers did see a significant improvement in both AUC and Fmax (CI) after finetuning (p=0.011 and p<0.001 respectively for the two measures). Given that there is a significant improvement in one of three cases, and slight improvement in the other two, our results weakly support the finetuning of large neural networks for CI prediction.   

\subsection{The utility of acoustic and linguistic features for CI prediction}
The use of both linguistic and acoustic features employed by the methods evaluated enabled us to also compare the two modalities for predicting CI status. Overall, the acoustic features were able to predict CI better than the linguistic ones (see Farzana et al’s unimodal configurations in Figure \ref{eval:farzana}, as well as W2V2 and IS10’s performance compared to BERT in Figure \ref{eval:ying}). Specifically, unimodal classifiers with interpretable acoustic features were among the top performers by both mean AUC and Fmax (CI). For instance, the IS10 feature set, used in Ying et. al.'s best-performing unimodal acoustic-based classifier, was designed to capture paralinguistic characteristics of speech, and were originally employed to predict age, gender, and affect \cite{schuller2010interspeech}. These features significantly outperformed the fine-tuned BERT features used by the method in terms of both AUC and Fmax (CI) (p<0.001 for both measures). In addition, the prosody features used in Farzana et. al.'s unimodal acoustic-based classifier capture suprasegmental features of speech, including how quickly an individual talks, how much silence is present in a recording, and how an individual's pitch varies while talking \cite{frick1985communicating}. These acoustic features significantly outperformed the expert-defined, interpretable NLP features used by the method in terms of both AUC and Fmax (CI) (p<0.001 for both measures). 

While interesting, this observation is counter to much of the literature on ADRD detection, where NLP-based methods tend to outperform acoustics-based methods \cite{luz2021alzheimer}. As a possible explanation, NLP features, such as use of pronouns and repetitions, have been found to vary significantly with ADRD in picture description tasks \cite{slegers2018connected}. As a broader category of impairment \cite{savva2015has}, CI may not give rise to these language deficits in the same way as ADRD, and thus may explain the divergence in trends.  

Another potential explanation for this divergence is that our cohort is more diverse than the ADReSSo-2021 dataset, and contains several patients who speak English as a second language. Recent studies have shown that acoustic-only models can predict MCI in a multilingual cohort, and that these methods outperform language-specific NLP methods \cite{agbavor2024multilingual}. It is possible that NLP features capturing syntactic structures of speech and word usage in English do not translate well to capturing CI in a diverse cohort. This is an exciting avenue for future research, as these interpretable acoustic features can be tested across multilingual cohorts much more easily than expert-defined NLP features that are specific to English.   

\section{Limitations and Conclusion of this Study}
A limitation of this study is the small size of the cohort. With only 28 patients assessed to have CI, it is difficult to assess how these methods will translate outside of this cohort. Furthermore, 157 transcripts and recordings are relatively few for training/finetuning DL-based methods for prediction. As more research is conducted assessing CI from spontaneous speech and more data are made available, we hope that stronger and clearer conclusions can be made in the future. 

Another limitation is that we only evaluated six open-source algorithms. Other non-open-source algorithms for the prediction of ADRD or MSA could have translated better to CI prediction than those that we have evaluated.  

Despite these limitations, we hope that our methods, findings and code \url{https://anonymous.4open.science/r/catch-C459} can provide an initial landscape of the predictability of CI from CTT recordings, and spur further progress in this critically needed area. Larger cohorts, more open-sourcing of prediction methods and the inclusion of other modalities, both derived from and in addition to CTT, are likely to contribute to this progress.

\bibliographystyle{IEEEtran}
\bibliography{custom}

\end{document}